%% file: main.tex
\documentclass[11pt,a4paper]{article}
\usepackage[table,dvipsnames]{xcolor} 
\usepackage[hyperref]{emnlp-ijcnlp-2019}
\usepackage{times}
\usepackage{latexsym}
\usepackage{adjustbox}

\usepackage{url}

\aclfinalcopy 

\usepackage{amsmath}
\usepackage{amsthm}
\usepackage{amsfonts}
\usepackage{color}
\usepackage{enumitem}
\usepackage{xspace}
\usepackage{booktabs,colortbl}
\usepackage{array}
\usepackage{etoolbox}

\input{macros}

\title{Discourse-Aware Semantic Self-Attention for Narrative Reading Comprehension}

\author{Todor Mihaylov \and Anette Frank \\
Research Training Group AIPHES\\
  Department of Computational Linguistics, Heidelberg University \\
  Heidelberg, Germany \\
  \tt\{mihaylov,frank\}@cl.uni-heidelberg.de \\\\
  }

\date{}

\begin{document}
\maketitle
\begin{abstract}
\input{sec_00_abstract.tex}

\end{abstract}

\input{sec_01_intro.tex}

\input{sec_02_annotations.tex}

\input{sec_03_approach_self_att.tex}

\input{sec_04_data.tex}

\input{sec_05_related.tex}

\input{sec_06_experiments.tex}

\input{sec_07_conclusion.tex}
\input{acknowledgements.tex}

\bibliography{main}
\bibliographystyle{acl_natbib}

\clearpage
\input{sup_appendix}

\end{document}

%% file: macros.tex
\newcommand\trainname{\textit{Train}\xspace}
\newcommand\devname{\textit{Dev}\xspace}
\newcommand\testname{\textit{Test}\xspace}

\usepackage[textsize=tiny, textwidth=2.5cm]{todonotes}
\usepackage{marginnote}
\newcounter{todonumber}
\newcommand{\note}[2][]{{%
 \let\marginpar\marginnote%
 \ifodd\value{todonumber}%
   \reversemarginpar%
 \else%
 \fi%
 \todo[#1]{#2}}%
 \stepcounter{todonumber}%
}

\newif\ifcameraready
\camerareadytrue

\ifcameraready
    
    \newcommand{\anette}[1]{#1}
    
    \newcommand{\revtwo}[1]{#1}
\else
    
    \newcommand{\anette}[1]{\textcolor{orange}{#1}}
    
    \newcommand{\revtwo}[1]{\textcolor{blue}{#1}}
\fi
\newcommand{\incorrect}[1]{\textcolor{BrickRed}{#1}}
\newcommand{\correct}[1]{\textcolor{OliveGreen}{#1}}

\newcommand{\goldanswer}[1]{\textcolor{RoyalBlue}{#1}}

\usepackage{color,soul}
\setul{0.5ex}{0.4ex}

\newcommand{\tc}[2]{\setulcolor{#1}\ul{#2}\setulcolor{black}}

\newcommand{\tcB}[1]{\tc{orange}{#1}}
\newcommand{\tcC}[1]{\tc{purple}{#1}}
\newcommand{\tcD}[1]{\tc{green}{#1}}

\newcommand{\tcA}[1]{\tc{cyan}{#1}}

\newcommand\rougeL{\textit{Rouge-L}\xspace}

\newcommand\bleuOne{\textit{Bleu-1}\xspace}

\newcommand\narrativeqa{NarrativeQA\xspace}

\newcommand\bidaf{BiDAF\xspace}
\newcommand\qanet{QANet\xspace}

\newcommand\Coref{Coref\xspace}
\newcommand\DR{DR\xspace}

\newcommand\discRelExp{DR (Exp)\xspace}
\newcommand\discRelNonExp{DR (NonE)\xspace}

\newcommand\SRL{SRL\xspace}
\newcommand\srl{SRL\xspace}
\newcommand\explicit{\textit{Explicit}\xspace}

\newcommand\srlDiscRelExp{SRL+\DR (Exp)\xspace}

\newcommand\srlDiscRelAllCoref{SRL + \DR (All) + Coref\xspace}

\newcommand{\STAB}[1]{\begin{tabular}{@{}c@{}}#1\end{tabular}}

\newcommand*{\img}[1]{%
    \raisebox{-.3\baselineskip}{%
        \includegraphics[
        height=\baselineskip,
        width=\baselineskip,
        keepaspectratio,
        ]{#1}%
    }%
}

%% file: sec_00_abstract.tex
In this work, we propose to use linguistic annotations as a basis for a \textit{Discourse-Aware Semantic Self-Attention} encoder that we employ for reading comprehension on long narrative texts.
We extract relations between discourse units, events and their arguments as well as coreferring mentions, using 
available annotation tools. 
Our empirical evaluation shows that the investigated structures improve the  overall performance (up to +3.4 \rougeL), especially intra-sentential and cross-sentential discourse relations, sentence-internal semantic role relations, and long-distance coreference relations. 
We show that dedicating self-attention heads to intra-sentential relations and relations connecting neighboring sentences is beneficial for finding answers to questions in longer contexts. Our findings encourage the use of discourse-semantic annotations to enhance the generalization capacity of self-attention models for reading comprehension.

%% file: sec_01_intro.tex
\section{Introduction}
\label{sec:intro}
Transformer-based self-attention models \cite{Vaswani2017-attention} have been shown to work well on many natural language tasks that require large-scale training data, such as Machine Translation \cite{Vaswani2017-attention,Dai2019-Transformer-XL}, Language Modeling \cite{radford2018-gpt1,Devlin2018Bert, Dai2019-Transformer-XL,radford2019language-gpt2} or Reading Comprehension \cite{QANet-Yu2018}, and can even be trained to perform surprisingly well in several multi-modal tasks \cite{Kaiser17-onemodel}. 
\input{figs/fig_motivation_example.tex}
\input{figs/fig_semantic_annotations.tex}

Recent work \cite{Strubell2018Lisa} has shown that for downstream semantic tasks with much smaller datasets,
such as Semantic Role Labeling (\SRL) \cite{palmer-etal-2005-srl}, self-attention models  greatly benefit from the use of linguistic information such as dependency parsing annotations.
Motivated by this work, we examine to what extent we can use discourse and semantic information to extend self-attention-based neural models for a higher-level task such as Reading Comprehension.

Reading Comprehension  is a task that requires a model to answer natural language questions, given a text as context: a paragraph or even full documents.
Many datasets have been proposed for the task, starting with a small multi-choice dataset \cite{Richardson2013-mctest-dataset}, large-scale automatically created cloze-style datasets \cite{Hermann2015-rc-cnn-dm,Hill2016-booktest} and big manually annotated datasets such as \citet{Onishi2016-rc-whodidwhat, Rajpurkar2016-squad, joshi-EtAl:2017:Trivia-qa,NarrativeQADeepMind2017}. Previous research has shown that some datasets are not challenging enough, as simple heuristics work well with them 
\cite{Chen2016-stanford-reader, Weissenborn2016-FastQa, Chen2016-reading-wikipedia-qa}. In this work we focus on the recent  \narrativeqa\cite{NarrativeQADeepMind2017} dataset that was designed not to be easy to answer and  that requires a model to read  narrative stories and answer questions about them. 

In terms of model architecture, previous work in reading comprehension and question answering has focused on integrating external knowledge (linguistic and/or knowledge-based) into recurrent neural network models 
using 
Graph Neural Networks \cite{Song2018-gnn}, Graph Convolutional Networks \cite{Sun2018-early-fusion, DeCao2018-gcn-coref}, attention \cite{Das2017-universal-schema,Mihaylov2018EnhanceCS, bauerwang2019commonsense} or pointers to coreferent mentions \cite{Dhingra2017-Linguistic-memory-RC}.

In contrast, in this work we examine the impact of \textit{discourse-semantic annotations} (Figure \ref{fig:motivation-example}) in a  \textit{self-attention architecture}. We 
build on
the QANet \cite{QANet-Yu2018} 
model 
by modifying the encoder of its self-attention modeling layer. 
In particular, we \textit{specialize self-attention heads to focus on
specific discourse-semantic annotations}, such as, e.g., an \textsc{arg1} relation in SRL, a \textsc{causation} relation holding between clauses in shallow discourse parsing, or coreference relations holding between entity mentions.

Our contributions are the following:
\begin{itemize}
    \item To our knowledge we are the first 
    to explicitly introduce discourse information into a neural model for reading comprehension. 
    
    \item We design a \textit{Discourse-Aware Semantic Self-Attention} mechanism, an extension to the standard self-attention models  -- without significant increase of computation complexity. 
    
    \item We analyze the impact of different discourse and semantic annotations for narrative reading comprehension \revtwo{and report improvements of up to 3.4 \rougeL over the base model}.

    \item We perform empirical fine-grained evaluation of the discourse-semantic annotations on specific question types and context size.
     
\end{itemize}

\revtwo{Code and data will be available at https://github.com/Heidelberg-NLP/discourse-aware-semantic-self-attention.} 

%% file: figs/fig_motivation_example.tex
\begin{figure}[t!]
\centering
\fbox{
\begin{minipage}{0.30\textwidth}
    \includegraphics[width=\textwidth]{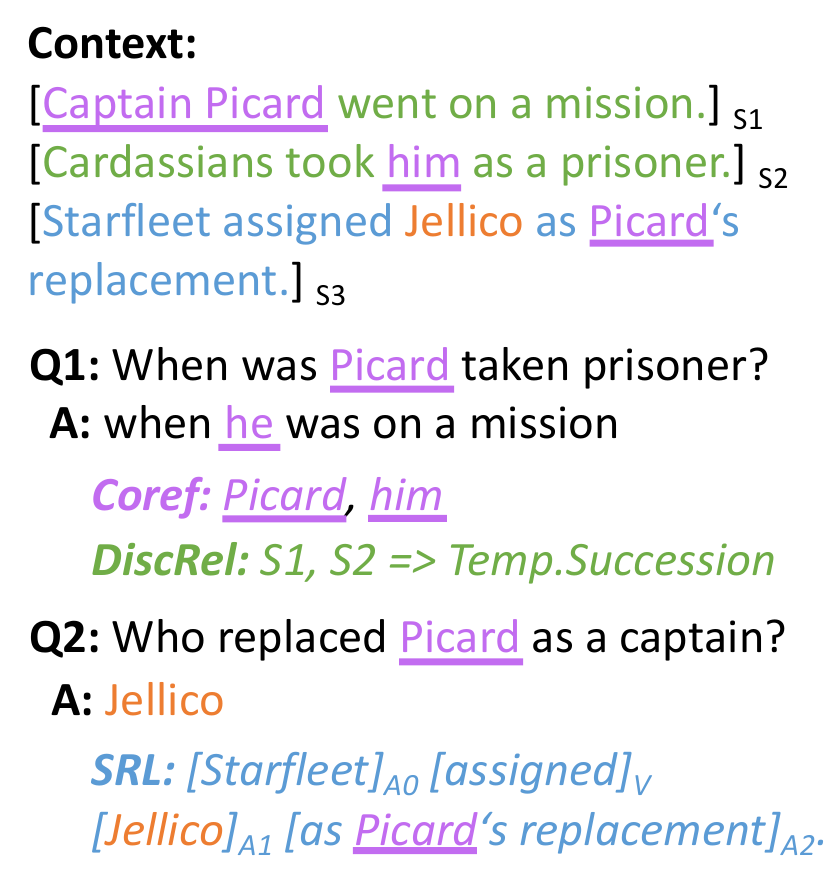}
\end{minipage}
}
  \caption{Motivational example: context and questions with required discourse and semantic annotations.}
  \label{fig:motivation-example}
\end{figure}

%% file: figs/fig_semantic_annotations.tex
\begin{figure*}[t!]
  \centering
  \includegraphics[width=0.95\textwidth]{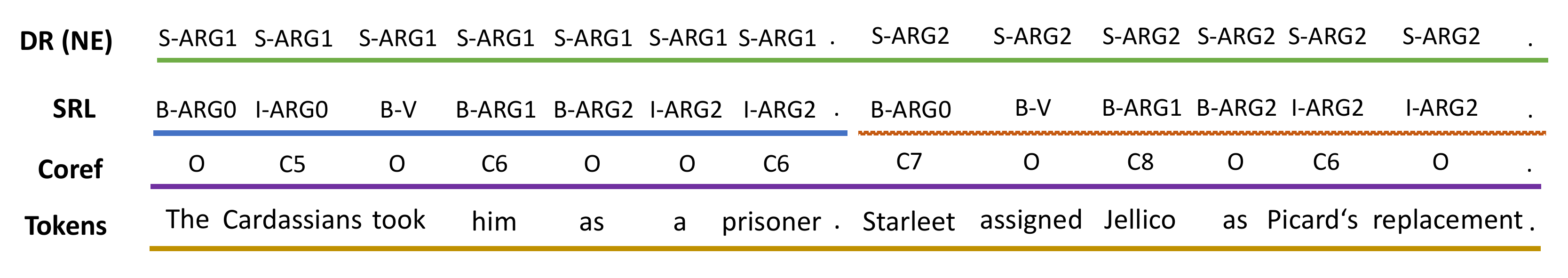}
  \caption{Example on different discourse-semantic annotations: DiscRel  (Dicourse Relations) (NE - Non-Explicit), SRL (Semantic Role Labeling), Coref (Co-reference resolution). The distinct horizontal lines show the interaction between the tokens: Coref - full context, SRL - single sentence, Non-Explicit DR - two neighbouring sentences.}
  \label{figure:annotation-example}
\end{figure*}

%% file: sec_02_annotations.tex
\section{Discourse-aware Semantic Annotations}
\label{sec:rc-with-semantics}
\input{figs/fig_03_semantic-encoder.tex}

Understanding narrative stories requires the ability to identify events and their participants and to identify how these events are related in discourse (e.g., by \textit{causation, contrast}, or \textit{temporal sequence}) \cite{mani2012-narrative}.
Our aim is to extract structured knowledge about these phenomena from long texts and to integrate this information in a neural self-attention model, in order to examine to what extent such knowledge can enhance the efficiency of a strong reading comprehension model applied to \narrativeqa.

Specifically, we enhance self-attention with knowledge about entity coreference (\Coref), their participation in events (\SRL) and the relation
between events in narrative discourse (Shallow Discourse Parsing \cite{xue-EtAl:2016:CoNLL-ST-SDP}, \DR).

All these linguistic information types are \textit{relational} in nature. 
For integrating relational knowledge into the self-attention mechanism, we follow a two-step approach: 
i) we extract such relations from a multi-sentence paragraph and \textit{project them
down to the token level}, specifically to the tokens of the text fragments that they involve; 
ii) we design a neural self-attention model that \textit{uses the interaction information between these tokens in a multi-head self-attention module}. 

To be able to map the extracted linguistic knowledge to paragraph tokens, we need annotations that are easy to map to token level (see Figure \ref{figure:annotation-example}). 
This can be achieved with tools for annotation of span-based 
Semantic Role Labeling, Coreference Resolution, and Shallow Discourse Parsing. 

\paragraph{Events and Their Participants } 
Relations between characters in a story are expressed in text through
their participation in states or actions in which they fill a particular event argument with a specific semantic role (see Figure \ref{figure:annotation-example}). 
For annotation of events and their participants we use the 
state-of-the-art 
SRL system of \citet{he-etal-2017-deep-srl} as implemented in AllenNLP \cite{Gardner2017AllenNLP}.
The system splits paragraphs into sentences and tokens, performs POS (part of speech tagging) and for each verb token V it predicts semantic tags such as ARG0, ARG1 (Argument Role 0, 1 of verb V), etc. 
When several argument-taking predicates are realized in a sentence, we obtain more than single semantic argument structure, and each token in the sentence can be involved in the argument structure of more than one verb. We refer to these annotations as different semantic views \cite{Khashabi2018-Semantic-views}, e.g., `semantic view for verb 1`. Different self-attention heads will be able to attend to individual semantic views.

\paragraph{Coreference Resolution}
Narrative texts abound of entity mentions that refer to the same entity in the discourse. 
We hypothesize that by directing the self-attention to this specific coreference  information, we can encourage the model to focus on tokens that refer to the same entity mention. Although token-based self-attention models are able to attend over wide-ranged context spans, we hypothesize that it will be beneficial to allow the model to focus directly on the parts of the text that refer to the same entity.
For coreference annotation we use the \textit{medium} size model from the neuralcoref spaCy extension available at https://github.com/huggingface/neuralcoref.
\revtwo{For each token we give as information the label of the corresponding coreference cluster (see Figure \ref{figure:annotation-example}) that it belongs to. Therefore, tokens from the same coreference cluster get the same label as input.}

\paragraph{Discourse Relations}
In narrative texts, events are connected by discourse relations such as \textit{causation, temporal succession}, etc. \cite{mani2012-narrative}. 
In this work we adopt the 15 fine-grained discourse relation sense types from the annotation scheme of the Penn Discourse Tree Bank (PDTB) \cite{Prasad-Dinesh-Lee-Miltsakaki-Robaldo-Joshi-Webber_LREC_PDTB}.  
For producing discourse relation annotations we use the discourse relation sense disambiguation system from \citet{MihaylovAndFrank2016-DR} which is trained on the data provided by the CoNLL Shared Task on Shallow Discourse Parsing \cite{xue-EtAl:2016:CoNLL-ST-SDP}. In this \revtwo{annotation} scheme discourse relations are divided into two main types: \textit{Explicit} and \textit{Non-Explicit}. \textit{Explicit} relations are usually connected with an explicit \textit{discourse connective}, such as \textit{because, but, if}.
\textit{Non-Explicit}\footnote{\textit{Non-Explicit} relations include \textit{Implicit}, \textit{AltLex} and \textit{EntRel} relation from PDTB. See \citet{xue-EtAl:2016:CoNLL-ST-SDP} for details.}
relations are not explicitly marked with a discourse connective and the arguments are usually contained in two consecutive sentences (see Figure \ref{figure:annotation-example}).
\revtwo{To extract explicit discourse relations we take into account only arguments that are in the same sentence. We consider as separate arguments (ARG1 and ARG2) text sequences that are on the left and right of an explicit discourse connective (CONN): ex. '[Jeff went home]$_{ARG1\_CCR}$ [because]$_{CONN}$ [he was hungry.]$_{ARG2\_CCR}$, where CCR is Contingency.Cause.Reason'.
To provide \textit{Non-Explicit} discourse relation sense annotations, we annotate every consecutive pair of sentences with a predicted discourse relation sense type.}

%% file: figs/fig_03_semantic-encoder.tex
\begin{figure*}[t!]
  \centering
  \includegraphics[width=0.97\textwidth]{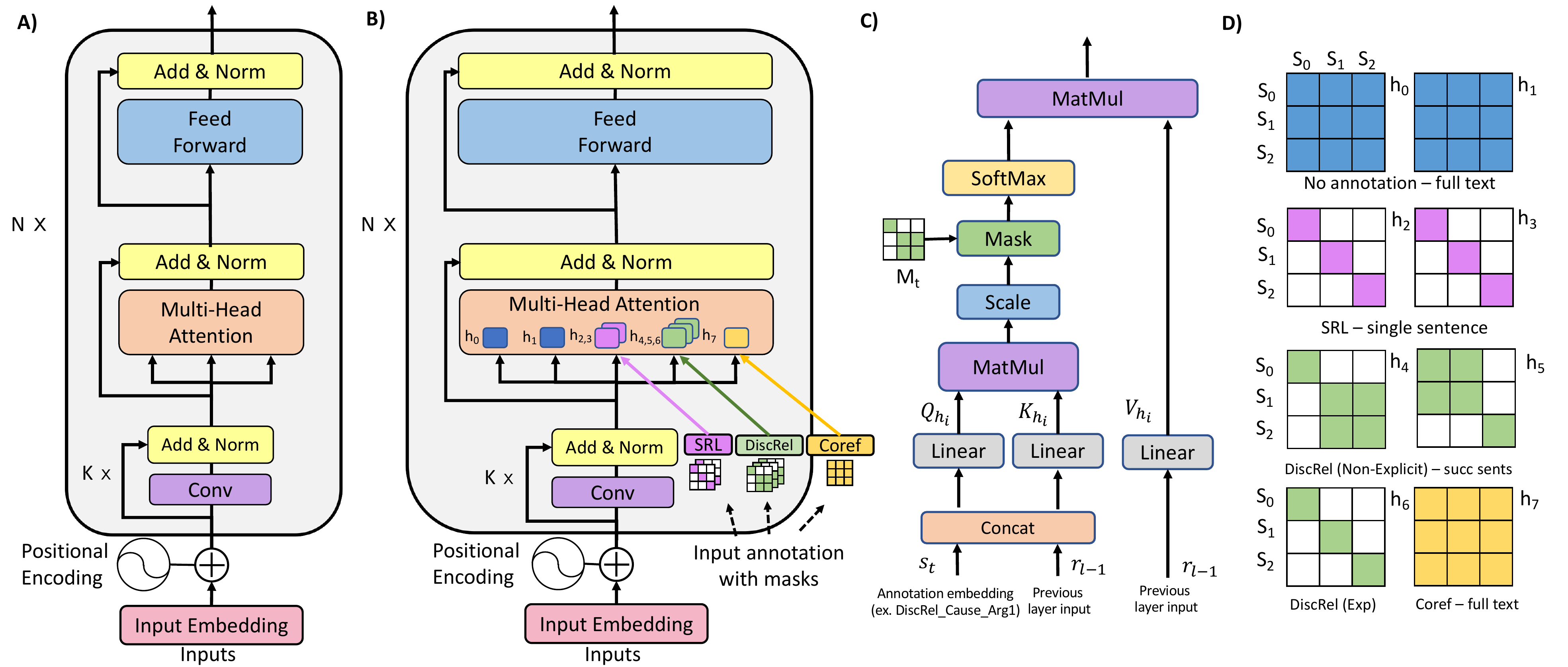}
  \caption{A) Base Multi-Head Self-Attention Encoder Block, B) Discourse-Aware Semantic Self-Attention (DASSA) Encoder Block , C) Single Attention Head with disource/semantic Information, D) Example of attention scope masks for different attention heads and different information.}
  \label{figure:self-attention}
\end{figure*}

%% file: sec_03_approach_self_att.tex
\section{A Discourse-Aware Semantic Self-Attention Neural Model}
\subsection{QANet}
As a base reading comprehension model we use QANet \cite{QANet-Yu2018}. QANet is a standard token-based self-attention model with
the following components, which are common across many recent models:
\textit{1.\ Input Embedding Layer} uses pre-trained word embeddings and convolutional character embeddings; \textit{2.\ Encoder Layer} consists of stacked \textit{Encoder Blocks} (see Figure \ref{figure:self-attention}, A) based on \textit{Multi-Head Self-Attention} \cite{Vaswani2017-attention} and \textit{depth-wise separable convolution}  \cite{Chollet16a-depthwise-cnn, KaiserGC17-depthwise}; \textit{3.\ Context-to-Query Attention Layer} is a standard layer, that builds a token-wise attention-weighted question-aware context representation; \textit{4.\ Modeling Layer} has the same structure as 2. above but uses as input the output of layer 3.; \textit{5.\ Output layer} is used for prediction of \textit{start} and \textit{end} answer pointers. For detailed information about these layers, please refer to \citet{QANet-Yu2018}. 
In this work we replace the standard \textit{Multi-Head Self-Attention} with \textit{Discourse-Aware Semantic Self-Attention}, using several different semantic and discourse annotation types. We describe this below and explain the differences to the standard \textit{Encoder Block}.

\subsection{Discourse-Aware Semantic Self-Attention}

In Figure \ref{figure:self-attention} we show the difference between the Base Multi-Head Self-Attention Encoder Block A) and the Discourse-Aware Semantic Self-Attention Encoder Block B). Both consist of $\operatorname{positional-encoding} + \operatorname{convolutional-layer} \times K + \operatorname{multi-head -self-attention} + \operatorname{feed-forward~layer}$. The difference is that B 
is provided additional inputs that are used by multi-head self-attention. The multi-head self-attention is a concatenation of outputs from multiple single self-attention heads $h_i$ followed by a linear layer.
A single head of the extended multi-head self-attention is shown in Figure \ref{figure:self-attention}C and is formally defined as
\begin{align}
a_{h_i}=\operatorname{mask\_softmax}\left(\frac{Q_{h_i} K_{h_i}^{T}}{\sqrt{d_{h}}}, M_{t}\right) V_{h_i} \\
Q_{h_i} = W^{Q}_{h_i}[r_{l-1};s_{t}] \in \mathbb{R}^{n \times d_{h}}\\
K_{h_i} = W^{K}_{h_i}[r_{l-1};s_{t}]  \in \mathbb{R}^{n \times d_{h}}\\
V_{h_i} = W^{V}_{h_i}r_{l-1} \in \mathbb{R}^{n \times d_{h}}
\end{align},
where $Q_{h_i}$, $K_{h_i}$, $V_{h_i}$ are components of the query-key-value attention and $\sqrt{d_{h}}$ is used for weight scaling as originally proposed in \citet{Vaswani2017-attention}. 
$W^{Q}_{h_i}$, $W^{K}_{h_i}$, $W^{V}_{h_i}$ are weights, specific for head $h_{i}, i \in 1..H$ \footnote{Number of heads H=8 as in original QANet specification if not specified otherwise.}, $r_{l-1}$ is the input from the previous encoder block, $s_{t}$ is an embedding vector for the linguistic annotation type $t$ (`SRL\_Arg1`, `DiscRel\_Cause.Reason\_Arg2`, etc.),
$a_{h_i}$ is the output of head $h_i$. $M_{t}$ is a sentence-wise attention mask as shown in Figure \ref{figure:self-attention}D. 
$s_t$ and $M_{t}$ are the \revtwo{main difference compared to the standard self-attention} (Figure \ref{figure:self-attention}C). 

In principle, representing edges of a graph (e.g., the V-ARG1 
role from SRL) requires memory of 
\revtwo{$n^2d_{h}H$, where $n$ is the length of the context}, which would be a bottleneck for computation on a GPU with limited memory (8-16GB). Instead, we adopt a strategy where the \textit{relation} is represented as a \textit{source and target node} and an \textit{attention scope} (one sentence for SRL; two sentences for DR (Non-Exp); full context for Coref). The latter is controlled using the attention mask. The combination of flat token labels and mask reduces the maximum memory required for representing the information in 
the knowledge-enhanced head to \revtwo{$2nd_{h}H$}.
The attention masks, which we use for reducing the attention scope of the different semantic and discourse annotations, are shown in Figure \ref{figure:self-attention}D. These masks ensure that the corresponding attention heads will only attend to tokens from the corresponding scope (SRL: single sentence; \discRelNonExp: two sentences, etc.). The attention masks are symmetric to the matrix diagonal. 
Therefore, they can easily be computed `on-the-fly` given only the sentence boundaries (corresponding to the 
horizontal lines in Figure \ref{figure:annotation-example}).

\revtwo{To reduce the model memory further and still benefit from the full-context self-attention, we use the \textit{Discourse-Aware Semantic Self-Attention} encoder (Figure \ref{figure:self-attention}B) only for blocks [1,3,5] of the \textit{Modeling Layer} that consists of 7 stacked encoder blocks (indexed 0 to 6). Blocks [0,2,6] are set as the base encoders that look at the entire context (Figure \ref{figure:self-attention}A).}

%% file: sec_04_data.tex
\section{Data and Task Description}
\label{sec:data}
\paragraph{NarrativeQA}
We perform experiments with the NarrativeQA
\cite{NarrativeQADeepMind2017} 
reading comprehension dataset. This dataset requires understanding of narrative stories (English) in order to provide answers for a given question. It
offers
two sub-tasks: (i) answering questions about a long narrative summary 
(up to 1150 tokens)
of a book or movie, or (ii) answering questions about entire books or movie scripts of lengths
up to 110k tokens. 
We are focusing on the summary setting (i) and refer to the summary as \textit{document} or \textit{context}. 
The dataset contains 1572 documents in total, devided into \trainname (1102 docs, 32.7k questions), \devname (115 documents, 3.5k questions) and \testname (355 documents, 10.5k questions) sets. 

\paragraph{Generative QA as Span Prediction}
An interesting aspect of the NarrativeQA dataset is that in contrast to most other RC datasets, the two answers provided for each question are written by human annotators. 
Therefore, answers typically differ in form from the context passages 
that license them.
To map the human-generated answers to answer candidate spans from the context, we use \rougeL \cite{lin-2004-rouge} to calculate a  similarity score between token n-grams from the provided answer and token n-grams from candidate answers selected from the context (we select candidate spans of the same length as the given answer). If two answer candidates have the same \rougeL score, we calculate the score between the candidates' surrounding tokens (window size: 15 tokens to the left and  right) and the question tokens, and choose the candidate with the higher score. We retrieve the best candidate answer span for each answer and use the candidate with the higher \rougeL score as supervision for training. 
We refer to this method for answer retrieval as \textit{Oracle (Ours)}.

%% file: sec_05_related.tex
\section{Related Work}
\label{sec:related-work}


\paragraph{Reading Comprehension with Knowledge}
Recent work has proposed different approaches for integrating external knowledge into neural models for the
high-level downstream tasks 
reading comprehension 
(RC) 
and question answering 
(QA). One 
line of work leverages external knowledge from knowledge bases for RC \cite{xu-etal-2016-question,Weissenborn17-knowledge,MCScript2018Ostermann,
Mihaylov2018EnhanceCS, bauerwang2019commonsense, Wang2018-SemevalCN} 
and QA \cite{Das2017-universal-schema,Sun2018-early-fusion, Tandon2018ActionsAndCommonsense}. These approaches make use of
implicit \cite{Weissenborn17-knowledge} or explicit \cite{Mihaylov2018EnhanceCS,Sun2018-early-fusion, bauerwang2019commonsense} attention-based knowledge aggregation or leverage
features from knowledge base relations \cite{Wang2018-SemevalCN}. Another 
line of work builds on linguistic knowledge from downstream tasks, such as coreference resolution \cite{Dhingra2017-Linguistic-memory-RC} or notions of co-occurring candidate mentions \cite{DeCao2018-gcn-coref} and OpenIE triples \cite{Khot2017QAOpenIE} 
into RNN-based encoders.
Recently, several pre-trained language models \cite{elmo-Peters:2018,Radford2018OpenaiGPT,Devlin2018Bert} have been shown to incrementally boost the performance of well-performing models for several short paragraph reading comprehension tasks \cite{elmo-Peters:2018,Devlin2018Bert} and question answering \cite{Sun2018-ReadingStrategies}, as well as many tasks from the GLUE benchmark \cite{Wang2018GLEUBenchmark}. 
Approaches based on BERT \cite{Devlin2018Bert} usually perform best when the weights are fine-tuned for the specific training task.
Earlier, many papers that do not use self-attention models or even neural methods
have also tried to use semantic parse labels \cite{yih-etal-2016-value}, or annotations from upstream tasks \cite{Khashabi2018-Semantic}.


\paragraph{Self-Attention Models in NLP} 
Vanilla self-attention models \cite{Vaswani2017-attention} use positional encoding, sometimes combined with local convolutions \cite{QANet-Yu2018} to model the token order in text. 
Although they are scalable due to their recurrence-free nature, most self-attention models do not well work when trained with fixed-length context, due to the fact that they often learn global token positions observed during training, rather than relative. To address this issue, \citet{Shaw2018-relative-position} proposes relative position encoding to model the distance between tokens in the context. \citet{Dai2019-Transformer-XL} address the problem of moving beyond fixed-length context by adding recurrence to the self-attention model.
\citet{Dai2019-Transformer-XL} argue that the fixed-length segments used for language modeling hurt the performance due to the fact that they do not respect sentence or any other semantic boundaries. 
In this work we also support the claim that the lack of semantic, and also discourse \textit{boundaries} is an issue, and therefore we aim to introduce structured linguistic information  into the self-attention model.
We hypothesize that the lack of local discourse context is a problem for answering narrative questions, where the answer is contained inside the same sentence, or neighbouring sentences and therefore, by offering discourse-level semantic structure to the attention heads, offer ways to restrict, or focus the model to wider or narrower structures, depending on what is needed for specific questions.

Self-attention architectures can be seen as graph architectures (imagine the token (node) interactions as adjacency matrix) and 
are applied to
graph problems \cite{velickovic2018graph-att,li2019graph-transformer}. 
Therefore, in very recent work \citet{Kedziorski2019-Lapata-graph2text-transformer} have used a self-attention encoder as a graph encoder for text generation, in a dual encoder model. 
A dual-encoder model similar to \citet{Kedziorski2019-Lapata-graph2text-transformer} is suitable for a setting  where the input is knowledge from a graph knowledge-base. For a text-based setting like ours, where word order is important and the tokens are part of semantic arguments, an approach that tries to encode linguistic information in the same architecture \cite{Strubell2018Lisa} 
is more appropriate. 
Therefore our method is most related to LISA \cite{Strubell2018Lisa}, which uses joint multi-task learning of POS and Dependency Parsing to inject 
syntactic information for Semantic Role Labeling. In contrast, we do not use multi-task learning, but 
directly encode
semantic information extracted by pre-processing with existing tools.

\input{tables/table-01-overall-results.tex}
\input{tables/table-attention-heads.tex}
\paragraph{NarrativeQA} The summary setting of the NarrativeQA dataset \cite{NarrativeQADeepMind2017} has in the past been addressed with attention mechanisms by the following models:
\textit{BiAtt + MRU} 
\cite{MultiRange-Tay2018} is similar to \textit{BiDAF} \cite{Seo2017-bidaf}. 
It is bi-attentive (attends form context-to-query and vice versa) but enhanced with a MRU (Multi-Range Reasoning Units).
MRU is a compositional encoder that splits the context tokens into ranges (n-grams) of different sizes and combines them in summed n-gram representations and fully-connected layers.
\textit{DecaProp }\cite{Tay2018-Densely-connected} is a neural architecture for reading comprehension, that densely connects all pairwise layers, modeling relationships between passage and query across all hierarchical levels. \citet{bauerwang2019commonsense} observed that some of the questions require external commonsense knowledge and developed \textit{MHPGM-NOIC} - a \textit{seq2seq} generative model with a copy mechanism that also uses commonsense knowledge and ELMo \cite{elmo-Peters:2018} contextual representations.
\citet{hu-etal-2018-attention-distill} used an implementation of Reinforced Mnemonic Reader (\textit{RMR}) \cite{HuPQ17-RMR}. They also proposed \textit{RMR + A2D}, a novel teacher-student attention distillation method to train a model to mirror the behavior of the ensemble model \textit{RMR (Ens)}.

%% file: tables/table-01-overall-results.tex
\begin{table}[t!]
\small
\centering
\begin{tabular}{lll}
\hline
Model                                         & B-1 & R-L \\\hline
\multicolumn{3}{c}{\textbf{\cite{NarrativeQADeepMind2017}}} \\\hline
Human                                         & 44.43  & 57.02   \\
Oracle (original) & 54.14  & 59.92   \\
Seq2seq (no context) \dag                         & 15.89   & 13.15   \\
ASR  \dag                                          & 23.30   & 22.26   \\
BiDAF                                         & 33.72  & 36.30    \\\hline
\multicolumn{3}{c}{\textbf{Previous work}} \\\hline
BiAtt + MRU \cite{MultiRange-Tay2018}                           & 36.55  & 41.44   \\
DecaProp \cite{Tay2018-Densely-connected}    & 44.35  & 44.69   \\
MHPGM + NOIC \cite{bauerwang2019commonsense} \dag                                 & 43.63  & 44.16   \\
RMR \cite{hu-etal-2018-attention-distill}     & 48.40   & 51.50     \\\hline
RMR (Ens) \cite{hu-etal-2018-attention-distill}                                       & 50.10   & 53.90    \\
RMR + A2D \cite{hu-etal-2018-attention-distill}                                     & 50.40  & 53.30  \\\hline  
\multicolumn{3}{c}{\textbf{This work}} \\\hline
Oracle (ours)                                 & 70.71  & 70.82   \\
BiDAF                                         & 47.19   & 49.63   \\
QANet                                         & 46.37   & 48.66   \\
+ DR (Exp)                         & 50.12   & 52.14\\
+ DR (Exp) EMA                        & 51.16   & 53.26
\\\hline
\end{tabular}
\caption{Results on the \narrativeqa \testname set. Models with \dag~are generative, while the rest use span prediction.}
\label{table:overal-results}
\end{table}

%% file: tables/table-attention-heads.tex
\begin{table}[ht!]
\small
\centering
\begin{tabular}{llllll}\\\hline
Config                      & \STAB{\rotatebox[origin=c]{90}{DR-E}} & \STAB{\rotatebox[origin=c]{90}{DR-NE}} & \STAB{\rotatebox[origin=c]{90}{SRL}} & \STAB{\rotatebox[origin=c]{90}{Coref}} & \STAB{\rotatebox[origin=c]{90}{No}} \\\hline
QANet (baseline)           & -   & -    & -   & -     & 8  \\\hline
DR (All)               & 2   & 2    & -   & -     & 4  \\
DR (Exp)               & 2   & -    & -   & -     & 6  \\
DR (NonE)              & -   & 2    & -   & -     & 6  \\
Coref              & -   & -    & -   & 3     & 5  \\
SRL                         & -   & -    & 3   & -     & 5  \\
SRL+ DR (Exp)         & 2   & -    & 3   & -     & 3  \\
SRL + DR (NonE)        & -   & 2    & 3   & -     & 3  \\
SRL + DR (All)         & 2   & 2    & 3   & -     & 1  \\
SRL + DR (Exp) + Coref & 2   & -    & 3   & 1     & 2  \\
SRL + DR (All) + Coref & 2   & 2    & 3   & 1     & 4 \\\hline
\end{tabular}
\caption{The number of attention heads by discourse-semantic type. \textit{`No'} 
means that no linguistic annotation types are provided 
(attends to all tokens). }
\label{tab:config-attention-heads}
\end{table}

%% file: sec_06_experiments.tex
\section{Experiments and Results}
\label{sec:experiments}

In this section we describe the experiments and results of our proposed model in different configurations.
We compare the results of different models using overall results (Table \ref{table:overal-results}) on the dataset, but also the performance for different question types (Figure \ref{figure:question-types-qanet}) and context 
sizes (Figure \ref{figure:context-length}).

\subsection{Overall Results}
\input{figs/fig_01_question_type_improvement.tex}
Table \ref{table:overal-results} compares our baselines and proposed model to prior work. We report results for \bleuOne,  and \rougeL scores.
The first section lists results on the \narrativeqa dataset as reported in \textbf{\citet{NarrativeQADeepMind2017}}. 
\textit{Oracle (original)} uses the gold answers as queries to match a token sequence (with the answer length) in the context that has the highest \rougeL. 
\revtwo{In contrast, using \textit{Oracle (Ours)}, described in Section \ref{sec:data}, we report a +11 \rougeL score improvement (Table \ref{table:overal-results}: \textbf{This work}). }
The Oracle performance in this setting is important since the produced annotations are used for training 
of the span-prediction systems, and is considered upper-bound.\footnote{The previous work that uses span-prediction models do not report their \textit{Oracle} model used for training supervision.}
\textit{Seq2Seq (no context)} is an encoder-decoder RNN model trained only on the question.
\textit{ASR} is a version of the Attention Sum Reader \cite{Kadlec2016-as-reader} implemented as a pointer-generator that reads the question and points to words in the context that are contained in the answer.
\textit{BiDAF} is Bi-Directional Attention Flow \cite{Seo2017-bidaf} trained either with the \textit{Oracle (original)} or \textit{Oracle (ours)}.
The models from \textbf{Previous Work} are described in Section \ref{sec:related-work}.
In the last section of Table \ref{table:overal-results} we present the results of our experiments (\textbf{This work}). Here, \textit{BiDAF} and \textit{QANet} are implementations available in the AllenNLP framework \cite{Gardner2017AllenNLP}. In the last two rows we give the results of QANet extended with the proposed Discourse-Aware Semantic Self-Attention, 
using \revtwo{intra-sentential}, \textit{Explicit} discourse relations (\textit{DR (Exp)}, EMA is Exponential Moving Average). 

\subsection{Fine-grained Evaluation}
\label{sec:experiments-fine-grained}
\input{figs/fig_02_context_len.tex}
We further analyze
the performance of different configurations of our model 
by conducting 
fine-grained evaluation in view of question types (Figure \ref{figure:question-types-qanet}) and context length (Figure \ref{figure:context-length}). 

We define a range of system configurations using attention heads enhanced with different combinations of linguistic annotation
types,
including \textit{Explicit} (referred to as Exp or E) and \textit{Non-Explicit} (NonE, NE), \textit{Discourse Relations} (DiscRel, DR), \textit{Semantic Role Labeling} (SRL), and \textit{Coreference} (Coref), and configurations without any such additional information (\textit{No}). We also experiment with a setting where instead of using specific discourse relation types (such as DiscRel\_Exp\_Cause\_Arg1), we only identify that a token is part of \textbf{any} (NoSense) discourse relation (e.g., DiscRel\_Exp\_Arg1) or simply a multi-sentence attention span \textit{Sent span 3} with labels \textit{Sent1}, \textit{Sent2}, \textit{Sent3} for each sentence. This is to examine whether the type of discourse relation is important or rather the attention scope (intra-sentential, cross-sentence - 2, 3 neighbouring sentences, full context).

\paragraph{Question Type}
Different question types might profit from different linguistic annotation types. We thus
examine the performance of different question types, and analyze how it correlates with the presence of specific Semantic Self-Attention signals. 
We classify the questions into question types using a simple heuristic based on the question words as an indicator of their type (\textit{How / Where / Why / Who / What ...)}, and calculate the average \rougeL for each such questions type. 
The resulting scores are displayed in Figure \ref{figure:question-types-qanet}. 
In the first two columns of the figure, we report the \textit{Oracle} score and the baseline (\qanet) score. 
In the remaining columns we report (i) the improvement over the \revtwo{\qanet} baseline of \bidaf, and (ii) of our models with different combinations of discourse-aware semantic self-attention.
In the first row we report the score for each of the models on \textit{all} questions. 
We observe that best performing models on \textit{all} questions are the ones that include \explicit DR, and/or SRL. 
In terms of hardness, \textit{how} and \textit{why} questions usually have the lowest score. 
This not surprising since \textit{Oracle} performance is also low. For these type of questions, the RNN-based encoder (\bidaf) and self-attention with \discRelExp or \discRelNonExp perform best.
Almost all models with additional linguistic information improve over the baseline on \textit{when} questions, lead by the \srlDiscRelExp and \srlDiscRelAllCoref. 
\textit{What} questions are improved most by \discRelExp and \srl alone or when combined.
\textit{Who} questions gain most from discourse relations and all models that contain \SRL.
\input{figs/fig_analysis_example_01.tex}
\input{figs/fig_analysis_example_02.tex}
\input{figs/fig_analysis_example_03.tex}
\paragraph{Context Length}
In Figure \ref{figure:context-length} we present the performance on documents of different lengths, in number of tokens. All presented models are trained on the examples from the \trainname set with context up to 800 tokens.
Again, the models \discRelExp and \srlDiscRelExp show clear improvement across all context lengths. It is clear that all models show improvement over length \textit{800-1000}. This supports our hypothesis that discourse information is required for generalizing to longer contexts. One reason is that some of the questions can be answered with a local context (one-two sentences) which are better represented given short discourse scope (one-three sentences) or long dependencies given coreference.

In the evaluation of multiple model configurations we notice that in some cases \anette{a} single discourse/semantic type (e.g. \discRelExp) performs better than \anette{in} combination with \anette{others} (e.g. \srlDiscRelExp). 
We hypothesize that the reason is that the linguistic annotations work well \anette{in combination with} free \textit{No} attention heads (see Table \ref{tab:config-attention-heads}). 
Currently, we place multiple annotations on the same \textit{Encoder Block} which reduces the number of \anette{free} attention heads. 
\anette{For instance,} for \srlDiscRelExp,
each knowledge-enhanced encoder block has 3 \srl + 2 \discRelExp + 3 No heads. In future work we plan to use different annotation heads per \textit{Encoder Block (EB)}: e.g., 
\textit{EB0} has 3 \SRL + 5 No; \textit{EB1} has 2 \discRelExp + 6 No; etc. 

\paragraph{Success and Failure Examples}
In Figures \ref{fig:analysis-example-01-srl-coref-good-dr-bad}, \ref{fig:analysis-example-02-all-good}, \ref{fig:analysis-example-03-coref-dr-good-srl-bad} we show examples of context\footnote{The part that contains the correct answer.} and questions, together with the answers from human annotators and some of the examined models.\footnote{For easier reading, we color the \goldanswer{gold}, \correct{correct}, and \incorrect{wrong} answers and underline the mentions of different characters.}
We provide a hypothetical rationale of what we would need to answer the question. \footnote{\img{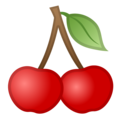} The examples are selected from \narrativeqa \testname, in such a way, that they depict the strength and weaknesses of the different models, corresponding to the empirical evaluation on Figure \ref{figure:question-types-qanet} and they fit in the space limit.}

%% file: figs/fig_01_question_type_improvement.tex
\begin{figure*}[t!]
  \centering
  \includegraphics[width=0.93\textwidth]{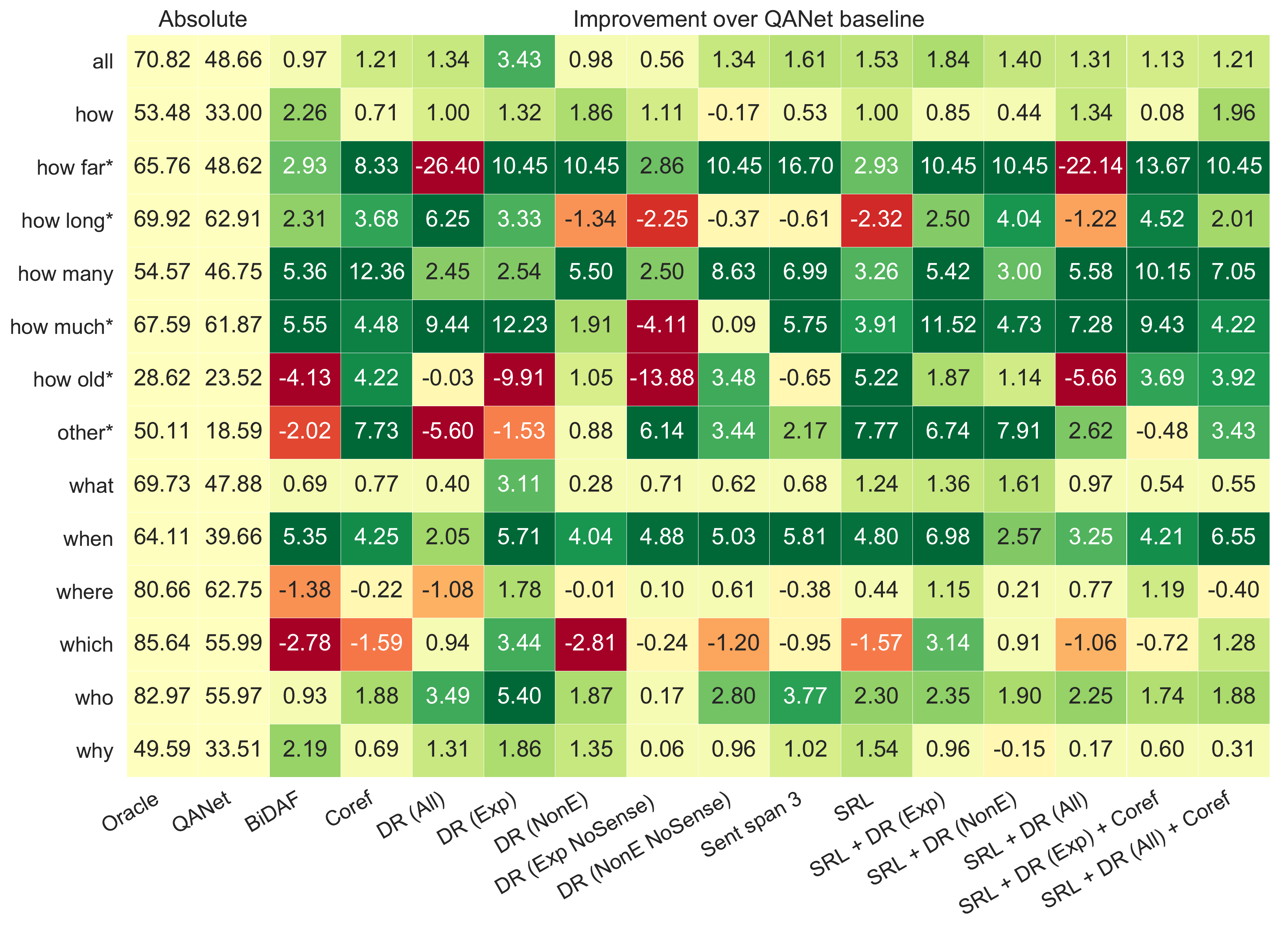}
  \caption{Rouge-L performance per Question Type on the \narrativeqa \testname set. The first two  columns represent \textit{Absolute} values. The rest are improvements  over the \qanet baseline model (i) by \bidaf and (ii) configurations of \qanet with linguistic information. Question types with * have less than 100 instances in the \testname set.}
  \label{figure:question-types-qanet}
\end{figure*}

%% file: figs/fig_02_context_len.tex
\begin{figure*}[t!]
  \centering
  \includegraphics[width=0.93\textwidth]{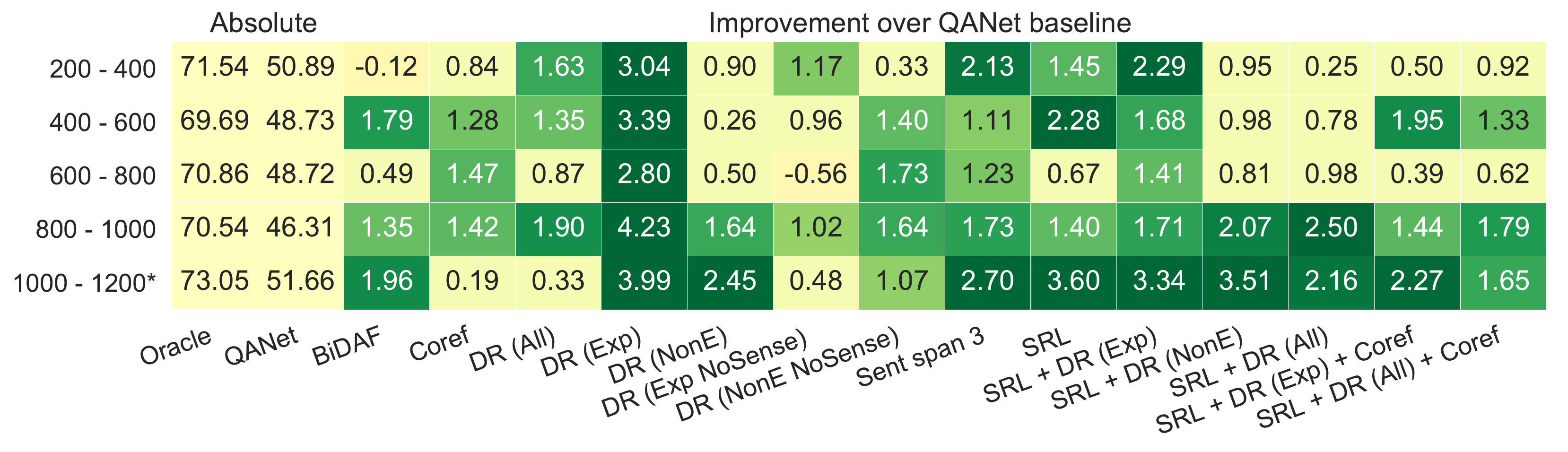}
  \caption{Rouge-L performance by context length on the \narrativeqa \testname set. The first two columns represent \textit{Absolute} values. The rest are improvements  over the \qanet baseline model (i) by \bidaf and (ii) different configurations of \qanet with linguistic information. Rows with * have less than 100 instances in the \testname set.}
  \label{figure:context-length}
\end{figure*}

%% file: figs/fig_analysis_example_01.tex
\begin{figure*}[t!]
\centering
\begin{minipage}{0.955\textwidth}
\small
\scalebox{0.95}{
\begin{tabular}{p{0.06\textwidth}p{0.904\textwidth}}
\hline
\textbf{Context} & Although \tcA{he} terrifies \tcB{the fairies} when \tcA{he} first arrives , \tcA{Peter} quickly gains favour with \tcA{them} . \correct{\tcA{He} amuses \tcB{them} with \tcA{his} human ways and agrees to play the panpipes at the \tcB{fairy} dances .} 
Eventually , Queen Mab grants \tcA{him} the wish of \tcA{his} heart , and \tcA{he} decides to return home to \tcA{his} mother . \\
\textbf{Question}  & After scaring the \tcB{fairies}, how does \tcA{Peter} win \tcB{them} over ? \\\hline
\multicolumn{2}{p{0.99\textwidth}}{
\textbf{Human 1:} \goldanswer{he agrees to play the panpipes at all of the fairy dances.};
\textbf{Human 2:} \goldanswer{He amuses them with his human ways and plays the pipes at their dances.};
\textbf{Oracle:} \correct{human ways and agrees to play the panpipes at the fairy dances} ;
\textbf{QANet:}  \incorrect{gains favour} ;
\textbf{DR (Exp), DR (NE):} \incorrect{quickly gains favour with them}; 
\textbf{Coref, SRL, SRL+DR(Exp):} \correct{He amuses them with his human ways and agrees to play the panpipes};
\textbf{SRL+DR(NE):} \correct{He amuses them with his human ways and agrees to play the panpipes at the fairy dances}
}
 \\\hline
 \multicolumn{2}{p{0.99\textwidth}}{
\textbf{Rationale:} To find the correct answer we need to know that (i) `gains favor' is a synonym to `win' in this context (commonsense); (ii) the following (2nd) sentence is the reason for the previous (1st) (DR - the model fails in this case) (iii) `\tcB{them}' are `\tcB{the fairies}', `\tcA{he}' is \tcA{Peter} (Coref)
}
 \\\hline
\end{tabular}
}
\end{minipage}
\caption{Example of positive impact of SRL and Coref and negative impact from discourse relations (DR).}
\label{fig:analysis-example-01-srl-coref-good-dr-bad}
\end{figure*}

%% file: figs/fig_analysis_example_02.tex
\begin{figure*}[t!]
\centering

\begin{minipage}{0.955\textwidth}
\small
\scalebox{0.95}{
\begin{tabular}{p{0.06\textwidth}p{0.904\textwidth}}
\hline
\textbf{Context} & \tcC{Jacob} frequently visits \tcA{Jeff }\tcB{and }\tcD{Kenny} , \tcA{w}\tcB{h}\tcD{o} are serving time in a juvenile hall . 
\correct{\tcC{Jacob} initially threatens \tcA{th}\tcB{e}\tcD{m}} , until eventually \tcA{Jeff} commits suicide . 
\tcC{Jacob} befriends \tcD{Kenny} , soon learning \tcD{he} has an early release and is illegally moving to New Mexico .\\
\textbf{Question}  & Why does \tcA{Jeff} committ suicide ? \\\hline
\multicolumn{2}{p{0.99\textwidth}}{
\textbf{Human 1:} \goldanswer{Jacob threatened them};
\textbf{Human 2:} \goldanswer{He is threatened by Jacob.};
\textbf{Oracle:} \incorrect{site which he says is} ;
\textbf{QANet:}  \incorrect{Jeff and Kenny , who are serving time in a juvenile hall};
\textbf{DR (Exp), DR (NE), SRL, SRL+DR(Exp), SRL+DR(NE):} \correct{Jacob initially threatens them ,}; \textbf{Coref:} \correct{Jacob initially threatens them} \incorrect{, until eventually Jeff commits suicide . Jacob befriends Kenny , soon learning he has an early release and is illegally moving to New Mexico}
}\\\hline
\multicolumn{2}{p{0.99\textwidth}}{
\textbf{Rationale:} To find the correct answer we need to understand that `until eventually' suggests that the suicide of \tcA{Jeff} is caused by \tcC{Jacob} threatening `\tcA{th}\tcB{e}\tcD{m}' (DR) and that \tcA{Jeff} is part of `\tcA{th}\tcB{e}\tcD{m}' (Coref).
}
 \\\hline
\end{tabular}
}
\end{minipage}

\caption{Example of positive impact of SRL and Coref, and discourse relations (DR).  
}
\label{fig:analysis-example-02-all-good}
\end{figure*}

%% file: figs/fig_analysis_example_03.tex
\begin{figure*}[t!]
\centering
\begin{minipage}{0.955\textwidth}
\small
\scalebox{0.95}{
\begin{tabular}{p{0.06\textwidth}p{0.904\textwidth}}
\hline
\textbf{Context} & 
The \tcA{four orphan children} of the house , \tcA{Edward , Humphrey , Alice and Edith} , are believed to have died in the flames . However , \tcA{they} are saved by \correct{\tcB{Jacob Armitage}} , a local verderer , \tcB{who} hides \tcA{them} in \tcB{his} isolated cottage and disguises \tcA{them} as \tcB{his} grandchildren .   Under \tcB{Armitage} 's guidance , \tcA{the children} from an aristocratic lifestyle to that of simple foresters . 
\\
\textbf{Question}  & Who rescues \tcA{the children} from fire at Arnwood ? \\\hline
\multicolumn{2}{p{0.99\textwidth}}{
\textbf{Human 1, Human 2:} \goldanswer{Jacob Armitage};
\textbf{Oracle:} \correct{Jacob Armitage}; 
\textbf{DR (Exp), DR (NE), Coref:} \correct{Jacob Armitage};
\textbf{QANet, SRL, SRL+DR(Exp):} \incorrect{Pablo};
\textbf{SRL+DR(NE):} \incorrect{Patience}
}
 \\\hline 
 \multicolumn{2}{p{0.99\textwidth}}{
\textbf{Rationale:} To find the correct answer we need to understand at least that `\tcA{they}' are `\tcA{the children}' (Coref) and `who did what to whom' in the context (SRL).
}\\\hline
\end{tabular}
}
\end{minipage}
\caption{Example of positive impact of Coref and DR and negative impact from SRL.}
\label{fig:analysis-example-03-coref-dr-good-srl-bad}
\end{figure*}

%% file: sec_07_conclusion.tex
\section{Conclusion and Future Work}
\label{sec:length}
In this work we use linguistic annotations as a basis for a \textit{Discourse-Aware Semantic Self-Attention} encoder that we employ for reading comprehension on narrative texts.
 
The provided annotations of discourse relations, events and their arguments as well as coreferring mentions, are using available annotation tools. 
Our empirical evaluation shows that discourse-semantic annotations combined with self-attention yields significant (+3.43 \rougeL) improvement over QANet's token-based self-attention when applied to NarrativeQA reading comprehension.
We analyzed the impact of different semantic annotation types on specific question types and context regions. \anette{We find, for instance, that} 
\SRL greatly improves \textit{who} and \textit{when} questions, and that discourse relations improve also the performance on \textit{why} and \textit{where} questions.
While all examined annotation types contribute,  particularly strong and constant gains are seen with intra-sentential DR (all context ranges), followed by SRL (short to mid-sized contexts). Coreference shows positive, but weaker impact, mostly in mid-sized contexts. 
\revtwo{A promising future direction would be to include additional external knowledge such as commonsense and world knowledge, and learn all annotations jointly with the downstream task.} 

%% file: acknowledgements.tex
\paragraph{Acknowledgments}
This work has been supported by the German Research Foundation as part
of the Research Training Group Adaptive Preparation of Information from Heterogeneous Sources (AIPHES) under grant No. GRK 1994/1. 
We thank NVIDIA Corporation for donating GPUs used in this research.

%% file: sup_appendix.tex
\appendix

\section{Appendix : Technical Details}
\label{sec:appendix}

Our code is implemented with AllenNLP \cite{Gardner2017AllenNLP} in Python 3. 
For BiDAF and QANet we use the default implementations from AllenNLP v0.8.2.
\footnote{BiDAF configuration  https://github.com/allenai/allennlp/\\blob/v0.8.2/training\_config/bidaf.jsonnet}
\footnote{QANet configuration  https://github.com/allenai/allennlp/\\blob/v0.8.2/training\_config/qanet.jsonnet}
For the QANet + Discourse-Aware Semantic Self-Attention we use the same hyper-parameters as QANet and we set the size of the label embeddings for the semantic information to $d_i=16$.

\subsection{Training Details}
We train all models  with Adam \cite{Kingma2015-adam} for up to 70 epochs with early stopping patience 20 and we halve the learning rate every 8 epochs if the Rouge-L on the \devname set does not improve.
Depending on the model size and the used GPU, we train the models with effective batch size of up to 32.
\footnote{When training a model with the desired batch size does not fit on a single GPU, we use accumulated gradients as explained at https://medium.com/huggingface/training-larger-batches-practical-tips-on-1-gpu-multi-gpu-distributed-setups-ec88c3e51255.}